\documentclass[letterpaper, 10 pt, conference]{ieeeconf} 

\IEEEoverridecommandlockouts                              
\usepackage{graphicx} 
\usepackage{times}
\usepackage{amsmath,amssymb}  
\usepackage{verbatim}
\usepackage{mathtools}
\usepackage{algpseudocode}
\usepackage[boxed]{algorithm2e}
\usepackage{tabularx} 
\usepackage{soul, color} 
\usepackage{float}
\usepackage{caption}
\usepackage{subfigure}

\usepackage[table,x11names]{xcolor}
\usepackage{url}
\usepackage{hyperref}
\SetKwRepeat{Do}{do}{while}
\usepackage{siunitx}
\usepackage{dsfont}
\usepackage{multirow}

\DeclareMathOperator*{\argmin}{argmin} 

\makeatletter
\def\endthebibliography{%
  \def\@noitemerr{\@latex@warning{Empty `thebibliography' environment}}%
  \endlist
}
\makeatother
\usepackage{amsmath}

\begin{document}

\title{\LARGE \bf Robot Guided Evacuation with Viewpoint Constraints}
\author{Gong Chen,  Malika Meghjani, Marcel Bartholomeus Prasetyo
    \thanks{The authors are with Singapore University of Technology and Design (SUTD), Singapore.
    {\tt\small chen\_gong@mymail.sutd.edu.sg}
    {\tt\small \{malika\_meghjani,marcel\_prasetyo\}@sutd.edu.sg}}
}
\maketitle

\begin{abstract}
We present a viewpoint-based non-linear Model Predictive Control (MPC) for evacuation guiding robots. Specifically, the proposed MPC algorithm enables evacuation guiding robots to track and guide cooperative human targets in emergency scenarios. Our algorithm accounts for the environment layout as well as distances between the robot and human target and distance to the goal location. A key challenge for evacuation guiding robot is the trade-off between its planned motion for leading the target toward a goal position and staying in the target's viewpoint while maintaining line-of-sight for guiding. We illustrate the effectiveness of our proposed evacuation guiding algorithm in both simulated and real-world environments with an Unmanned Aerial Vehicle (UAV) guiding a human. Our results suggest that using the contextual information from the environment for motion planning, increases the visibility of the guiding UAV to the human while achieving faster total evacuation time. 
\end{abstract}

\section{Introduction}

Search and rescue (SAR) operations are complex and often hazardous, requiring highly skilled professionals to work in challenging and unsafe environments. In recent years, robots have been increasingly used to assist human operators in SAR operations \cite{Murphy2016}, such as safe teleoperation \cite{goel2022Teleop}, lost target search \cite{9389479}, and victim identification \cite{lostpersonmodel2021Huintzman}. In this research, we focus on the evacuation aspect of SAR operations, specifically the problem of deploying robots to actively guide a human to the exit during emergencies. 

Robot-assisted evacuation is a highly challenging task that requires, effective communication \cite{convey_emergency_info2012Robin}, safe interaction between human and the robot \cite{safe_interaction_2011_PeterA} and fast adaptation to both the environmental factors and the human target's psycho-physiological factors \cite{psycho_physical_trust_emergency_2014_atkinson}.  
In this paper we address each of these challenges by proposing a viewpoint constrained guiding robot which leads a cooperative human target to the goal location. 
Unlike evacuation route exploration problems, which discover effective routes for humans to evacuate in unknown environments \cite{4543638}, our proposed algorithm searches for evacuation routes in known environment. However, we consider an additional constraint of actively guiding a target to a goal location. In addition to path exploration, maintaining the robot's visibility to the target is also critical for guiding the target. Robot's visibility to human target is studied in \cite{headPose2021Conte}, with a fixed position in the target's field of view. 
However, using our proposed robot guidance algorithm, the robot dynamically adapts its motion with respect to the target to consistently keep it in target's field of view. Thus, our proposed evacuation guiding algorithm based robot not only remains in view of the target providing implicit communication and safe interaction but also adaptively leads the target to the goal location.

Our proposed guidance behavior is defined using the target's viewpoint constraints, inter-agent distance costs, and geometric constraints of the environment through the use of Model Predictive Control (MPC). MPC is employed for local path planning to achieve effective guidance interaction with a cooperative target while avoiding obstacles using an offline map consisting of ellipsoids. Additionally, a global reference path is computed online to ensure that the evacuating pair can reach a desired exit.

\begin{figure}[t]
\centering
\includegraphics[width=\columnwidth]{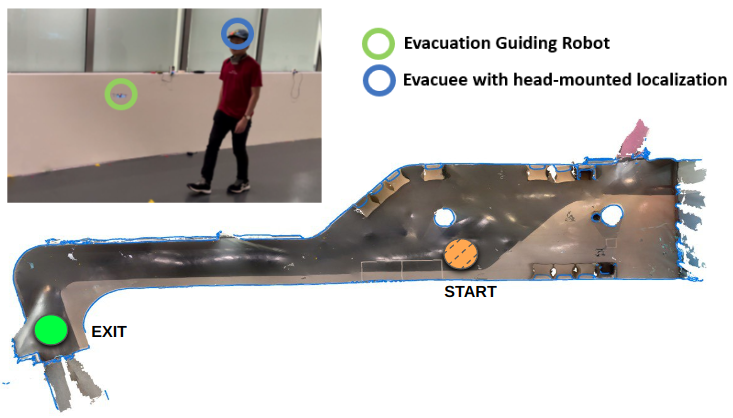}
\caption{\label{fig:floor_plan}UAV guiding a human follower in a L-shaped Corridor}

\vspace{-0.021\textheight}
\end{figure}

Our contributions in this paper are:
(a) a theoretical design of a robot-guided evacuation local planner using MPC, 
(b) a context-aware guidance local planner which uses prior knowledge about types of turns in a known environment and  
(c) a validation and demonstration of the proposed algorithm in simulated and real-world environments with a human evacuee.  
A real world setup of human following the evacuation guiding robot is illustrated in Fig \ref{fig:floor_plan}. 

\section{Related Work}\label{sec:related_work}

This paper focuses on the active guidance evacuation problem which involves planning a local path for an intelligent agent to guide a moving target to goal location. To tackle this problem, we divide it into three parts: (a) target tracking, (b) motion planning to achieve coordinated human-robot movement, and (c) desired evacuation guidance behavior.

Target tracking is often defined as a paparazzi problem \cite{895271}, where an agent captures an optimal view of a moving target that either moves away or stays in view of the agent intermittently. Rus et al. \cite{obs_avoidance_viewpoint_2017_Nageli} \cite{9636719} have studied this problem with a videography approach by defining camera shots or viewpoints from a robot's perspective to capture a moving target. 
Tobias et al. \cite{7847361} and Nanavati et al. \cite{dyad2019Amal} predict the intended trajectory of a human target by analyzing the target's head pose orientation. We define the viewpoint constraint based on the visual perspective of the human target. Also, we assume that the human target is cooperative, however, there are environmental uncertainties that may not allow the human to closely follow the robot.

For motion planning, Ray et al. \cite{aaron_thesis} proposed a mixed integer MPC model for a 2D agent for coordinated movement with a target. The mixed integer MPC model is used to define a series of viewpoint constraints, ensuring an agent's control output keeps its tracking target in view. Shkurti et al. \cite{florian} suggested a probabilistic target pursuit approach that uses topologically distinct paths for searching and following a moving target. Zhou et al. \cite{8967561} proposed a deep learning approach to solve for continuous control on a wheeled mobile robot to follow a leader. 
Our proposed algorithm builds on the non-linear MPC from \cite{aaron_thesis} and includes prior map information as contextual knowledge for the robot to have a higher probability of staying in the human target's view in obstacle-rich environments.

For the robot-guided evacuation behavior, Nayyar et al. \cite{8956307} suggest a multi-robot hands-off approach where stationary robots line up at checkpoints to gesture to a human target. Wagner \cite{robot_evac_paradigm} suggests a shepherding approach in theory where a single robot actively leads evacuees to goal location while ensuring the evacuees continue to follow after it. We formulate this approach as an active guidance behavior where the robot would alternate its local planner between maintaining line-of-sight with the human and leading the human toward a goal location. 
\section{Evacuation Guidance Algorithm}\label{sec:evacuation_algo}
\vspace{-0.005\textheight}
We propose a robot-guided evacuation algorithm that considers the viewpoint constraints from the target's perspective. The autonomous agent will constantly alternate its planned motion in either moving toward the target's viewpoint or leading the target toward a goal location. Furthermore, we consider prior contextual knowledge of the environment to reduce uncertainty in tracking and guiding it efficiently in a complex environment \cite{meghjani2019context}. 

\begin{figure}[hbt]
\centering
\includegraphics[trim={10 20 20 20},clip,width=0.5\textwidth]{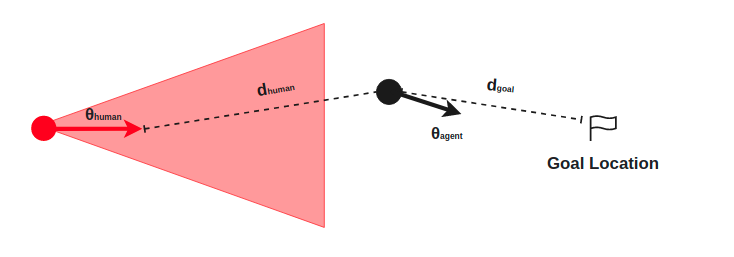}
\vspace{-0.010\textheight}

\caption{\label{fig:human_target}Example of agent guiding a human toward a goal}
\small
\end{figure}

The following is our problem formulation. An illustration of the evacuation scenario is presented in Fig. \ref{fig:human_target}, where the human target is represented as a red circle with a triangular field of view, the guiding agent is represented as the black circle, and the goal location, $\mathcal{X}_g$, is represented by the flag icon. 

The human and the agent are moving in directions $\theta_{human}$ and ${\theta}_{agent}$, respectively. Their positions in a 2D environment are represented by $\mathcal{X}_h$ and $\mathcal{X}_a$. We define the viewpoint of the human target as  $\mathcal{X}_h^*$, an orthogonal point $r$ distance away in the direction $\theta_{human}$: 
\begin{equation}
  \label{eq:view_point}
  \begin{aligned}
   \mathcal{X}_h^*= \begin{bmatrix}
        x_{human} + r*cos(\theta_{human})\\ y_{human} + r*sin(\theta_{human})
  \end{bmatrix}
  \end{aligned}
\end{equation}
The Euclidean distance between the human's viewpoint and the agent is denoted as $d_{human}$, where $d_{human} = \rVert \mathcal{X}_a - \mathcal{X}_h^* \rVert^2 $. The Euclidean distance between the agent and the goal location is denoted as $d_{goal}$, where  $d_{goal} =\rVert \mathcal{X}_a - \mathcal{X}_g \rVert^2 $. We assume the human target is cooperative and is willing to be evacuated.

For the proposed evacuation problem, we extend the work from \cite{aaron_thesis} where the target tracking problem is formulated as a nonlinear Model Predictive Control (MPC) problem. 
We introduce $\mathcal{X}_h^*$ as the estimated viewpoint position of the human target,  $\mathcal{X}_g$ as a predefined goal location, and $\mathcal{X}_{a}$ as the position of the agent. 
Given the above definitions, we formulate the agent's states by introducing cost functions to penalize the distance deviation from either the goal position or the target's viewpoint using distance functions $d_{goal}$ and $d_{human}$ respectively. 

To formulate the alternating behavior between moving toward the human and guiding human toward the goal position, we define $H$ as a balancing weight which we use to penalize on $d_{goal}$ and $d_{human}$. To ensure a smoother change in this weight, we transform $H$ with a Sigmoid function in Eq. \ref{eqn:h_balance}, enforcing it to be within a range of $(0,1)$, where $\beta$ and $c$ are constants.  $\beta$ controls how steeply the agent should change its motion between guiding and going back to the human's view. $c$ determines the minimum acceptable distance to be away from the target before leading.  
\begin{align}\label{eqn:h_balance}
H(\mathcal{X}_h^*,\mathcal{X}_a) =  \frac{\mathrm{1} }{\mathrm{1} + e^{- \beta * d_{human}(\mathcal{X}_h^*,\mathcal{X}_a) + c} } 
\end{align} The alternating behavior is achieved by multiplying the distance functions with $H$ and $(1-H)$, respectively. We define $C_{human}$ as the cost function to penalize the distance deviation between the agent and the target's viewpoint: 
\vspace{-0.01\textheight}
\vspace{-0.005\textheight}

\begin{equation}
  \label{eq:c_human}
  \begin{aligned}
        C_{human}(\mathcal{X}_h^*,\mathcal{X}_a)= H(\mathcal{X}_h^*,\mathcal{X}_a) * d_{human}(\mathcal{X}_h^*,\mathcal{X}_a)
  \end{aligned}
\end{equation}

The cost function $C_{goal}$ penalizes the distance deviation of the agent from the goal position. $T$ acts as a timestep-dependent constant that is calculated as: $(1-\gamma^k)$, where $\gamma$ is empirically defined as $0.95$ and $k$ represents the $k$-th timestep in the planning horizon. Thus, as $T$ increases, the agent would prioritize the penalization for the goal position toward the end of the planning horizon, similar to \cite{meghjani2012multi}.

\vspace{-0.015\textheight}
\begin{equation}
  \label{eq:c_goal}
  \begin{gathered}
        C_{goal}(\mathcal{X}_h^*,\mathcal{X}_a,\mathcal{X}_g) = (1- H(\mathcal{X}_h^*,\mathcal{X}_a)+T)* d_{goal}(\mathcal{X}_a,\mathcal{X}_g)
  \end{gathered}
\end{equation}

\begin{figure}[htb] 
\centering
\includegraphics[width=0.45\textwidth]{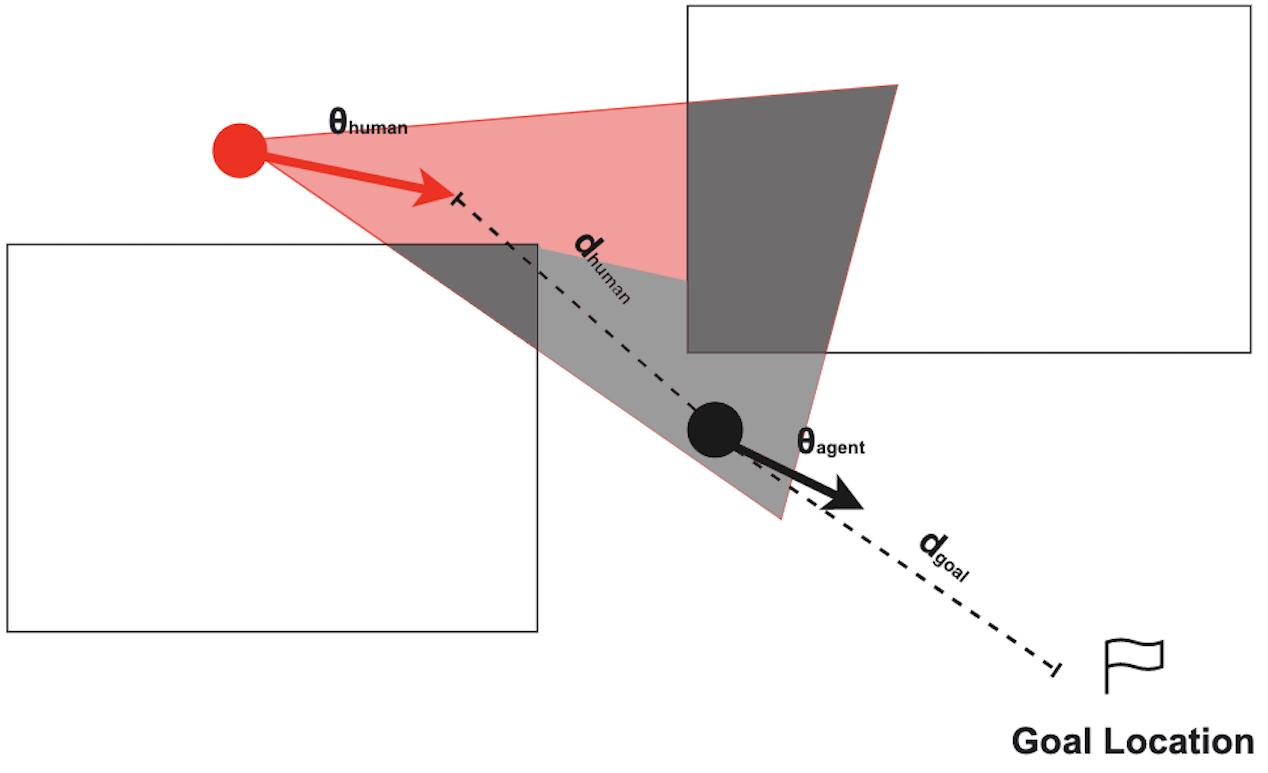}
\caption{\label{fig:context_undesired}Example with occluded target's view while turning.}
\end{figure}

Furthermore, we hypothesize that the prior knowledge of the environment helps the guiding agent in deviating less from the target's viewpoint. This is particularly important when there is a sharp turn in the evacuation route as certain field of view will be occluded by the obstacles. An undesired example of such occlusion can be seen in Fig.\ref{fig:context_undesired}.

Thus, we propose a cost function, $C_{context}$, for the agent to move closer toward the human target's viewpoint in the presence of turning points. $P_{context}^i$ accounts for any turning point, $i$, sensed in a radial range $R$ on the agent and is represented in Eq. \ref{eqn:p_value}
where, a turn could be \{"quarter", "T-turn", "cross intersection", "others"\}. For any turn that has only one path option, it will be classified as a "quarter" turn, and any turn with more than three path options will fall under "others". We assume that with a higher number of path options at each turning point, the agent should bias more toward the target's viewpoint. The reason for such an assumption is that the human target may take more time in understanding the intended direction of the leading agent when there are more path options. 
\vspace{-0.005\textheight}
\begin{equation}\label{eqn:p_value}
     P_{context}^i= 
\begin{cases}
     0& \text{if no turns} \\
     p_1              & \text{if quarter turn}\\
     p_2& \text{if T-turn}\\
     p_3& \text{if cross intersection}\\
     p_4& \text{any other turns}\\
\end{cases}
\end{equation}
\vspace{-0.005\textheight}

The values for $p_1$, $p_2$, $p_3$ and $p_4$ are non-negative. We sum all the $P_{context}^i$ sensed in the radial range $R$ at the position, $\mathcal{X}_a$, and use it as weighted constant to penalize $d_{human}$ to obtain $C_{context}$ in Eq. \ref{eq:c_context}. This allows us to regulate the motion of the agent such that it will prioritize remaining in the human target's viewpoint.
\vspace{-0.01\textheight}
\begin{equation}
  \label{eq:c_context}
  \begin{gathered}
        C_{context}(\mathcal{X}_h^*,\mathcal{X}_a) =  
   d_{human}(\mathcal{X}_h^*,\mathcal{X}_a) * \sum_{i \in R}^{} P_{context}^i 
  \end{gathered}
\end{equation}
 \vspace{-0.03\textheight}

\subsection{MPC Formulation}
 \vspace{-0.005\textheight}
Our objective function is to minimize the combined distance cost functions considering the target's viewpoint position, $\mathcal{X}_h^*$, the guiding agent's position at time $k$, $\mathcal{X}_{a_k}$, and the goal location, $\mathcal{X}_g$, in the planning horizon, $N$, as presented in Eq. \ref{eqn:mpc} below.
\begin{IEEEeqnarray}{rCl}\label{eqn:mpc}
    \argmin_{\boldsymbol{\mathcal{X}_{a_{1:N}},u_{k}}} \sum_{k = 1}^{N} &&  C_{human}(\mathcal{X}_h^*,\mathcal{X}_{a_{k}})  +C_{goal}(\mathcal{X}_h^*,\mathcal{X}_{a_{k}},\mathcal{X}_g)  \nonumber    \\&&+ C_{context}(\mathcal{X}_{a_{k}})+ g\lVert \boldsymbol{u_k} \rVert^2\IEEEyessubnumber \\
   \textrm{s.t.} &&  \boldsymbol{w}_1 = \boldsymbol{w}(0)\IEEEyessubnumber \\
   && \boldsymbol{u}(t_{c,k})= \frac{1}{2}(\boldsymbol{u}(t_{k})+\boldsymbol{u}(t_{k+1}))\IEEEyessubnumber\\
  && \boldsymbol{\dot w}(t_{c,k}) = f(\boldsymbol{w}(t_{c,k}),\boldsymbol{u}(t_{c,k}))\IEEEyessubnumber\\
  && \lVert(\mathcal{X}_{a_{k}}-m_i)^T M_i(\mathcal{X}_{a_{k}}-m_i)\rVert^2\leq d^i_{k}  \IEEEyessubnumber \\ 
  && \boldsymbol{w_k} \in W   \IEEEyessubnumber\\
  && \boldsymbol{u_k} \in U  \IEEEyessubnumber\\
  && \forall \; k \in {1,....,N}\IEEEyessubnumber
\end{IEEEeqnarray}

The agent's motion model follows a second-order unicycle model. The state vector $\boldsymbol{w}$, consists of the agent's position, 
heading direction, linear velocity, and angular velocity. It is denoted in order by:  $\boldsymbol{w} = [x \;\; y \;\; \theta \;\; v \;\;  \omega ]^T$. The control input vector, which consists of the agent's longitudinal and angular acceleration, is denoted by: $\boldsymbol{u} = [a \; \alpha]$. The dynamic model of the second-order unicycle model-based agent is defined by $ \boldsymbol{\dot w} = f(\boldsymbol{w},\boldsymbol{u}) = [v\cos(\theta) \;\; v\sin(\theta) \;\; \omega \;\; a \;\; \alpha]^T$. At each timestep $k$,
$w_k \in W \subset \mathbb{R}^{n_w}$ and $\boldsymbol{u_k} \in U \subset \mathbb{R}^{n_u}$, where $W$ and $U$ are the admissible range for the states and controls. Though the unicycle model is chosen for simplicity, other robot dynamic models can be similarly formulated given the same constraints as defined in the MPC formulation.

In Eq. (7a), we augment the control $\boldsymbol{u}$ into a square normal in the objective function, where $g$ is an arbitrary constant. This ensures a smoother change in control while optimizing for the distance costs. Eq. (7b) states the initial dynamic state of the agent. With reference to the direct collocation method described in \cite{Tedrake2022collocation}, the trajectory of the agent is sampled into collocation points, $t_{c,k}$. $\boldsymbol{u}$ is interpolated as a first-order polynomial in Eq. (7c). $\boldsymbol{w}$ is interpolated as a cubic polynomial over the interval $t \in [t_k,t_{k+1}]$ in Eq. (7d). The dynamics constraints are applied directly to $\boldsymbol{u}(t_{c,k})$ and $\boldsymbol{w}(t_{c,k})$.

To ensure collision avoidance, we refer to the \textbf{Proposition 1} presented in \cite{aaron_thesis}, where the obstacle-free space is computed offline as a graph of interconnected ellipsoids. At each timestamp, the agent's position, $\mathcal{X}_{a_{k}}$ is constrained within an obstacle-free ellipsoid with a center, $m_i$ and an orthogonal shape matrix $M_i$. At every timestep, only two connecting ellipsoids are considered for this constraint. The ellipsoid with the agent is denoted by $i = 0$. The next connecting ellipsoid as the agent moves in the prediction horizon is denoted by  $i = 1$. When $d^i_k < 1$, Eq. (7e) constrains the agent, $\mathcal{X}_{a_{k}}$ to be within the current and the next ellipsoids. For the remaining ellipsoids on the graph, $d^i_k$ is set to $\infty$, to relax the constraint.

\subsection{Global Path Planning} \label{para:global_path}
Given that the formulated MPC only considers adjacent ellipsoids in planning, the output trajectory only promises guidance toward the goal in the local region. If the goal position was located far in a complex environment, using local planning to reach the global goal becomes significantly ineffective. 
\begin{figure}[h]
\centering
\includegraphics[width=0.40\textwidth]{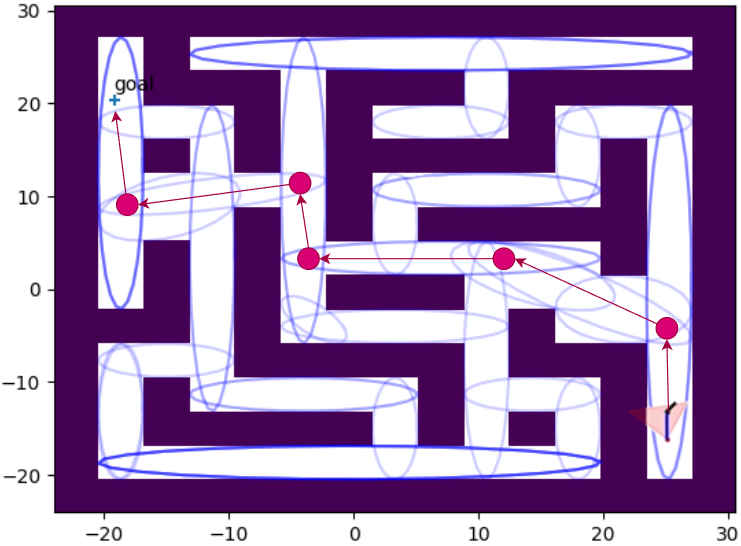}
\caption{\label{fig:sequential_goal}Example of sequential goals using ellipsoidal intersections.}
\end{figure}

\begin{figure*}[t!]
\centering
\subfigure[Simple Environment]{\includegraphics[trim={60 50 60 60},clip,width=0.64\columnwidth]{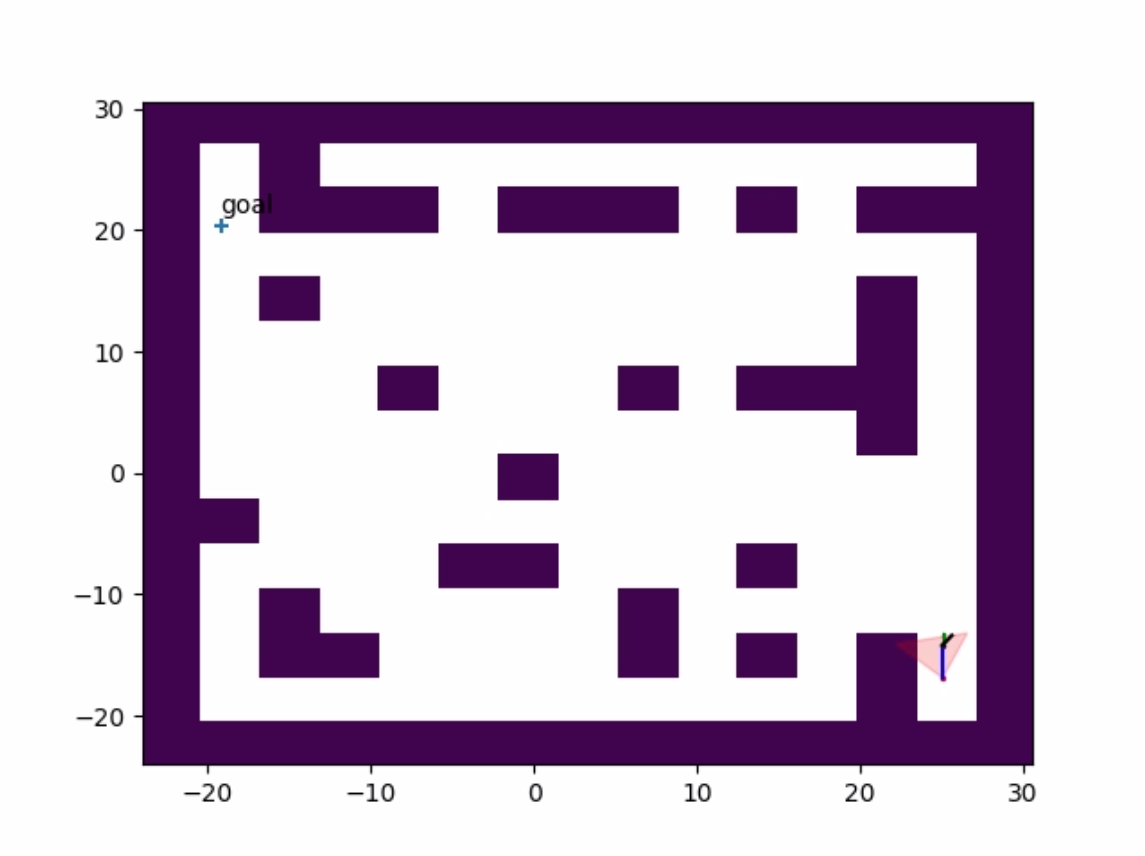}}
\subfigure[Maze Environment]{\includegraphics[trim={60 50 60 60},clip,width=0.64\columnwidth]{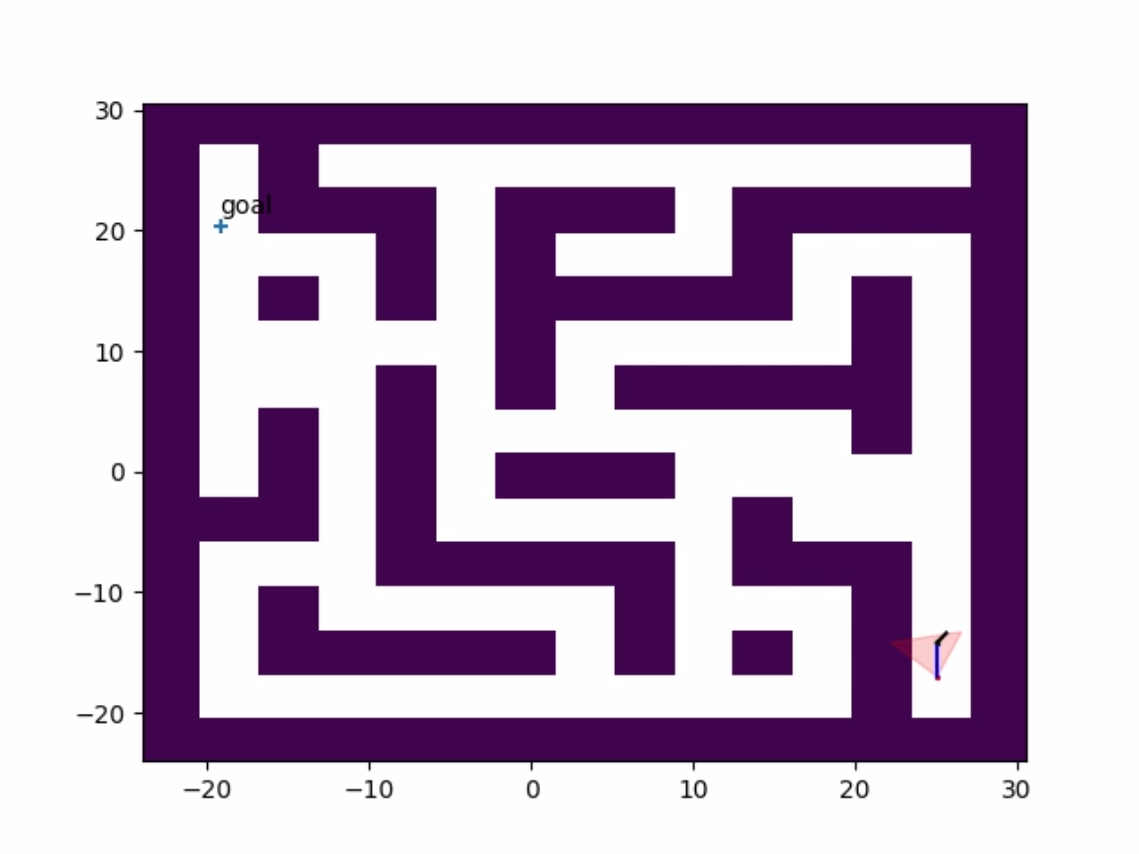}}
\subfigure[Forest Environment]{\includegraphics[trim={60 50 60 60},clip,width=0.64\columnwidth]{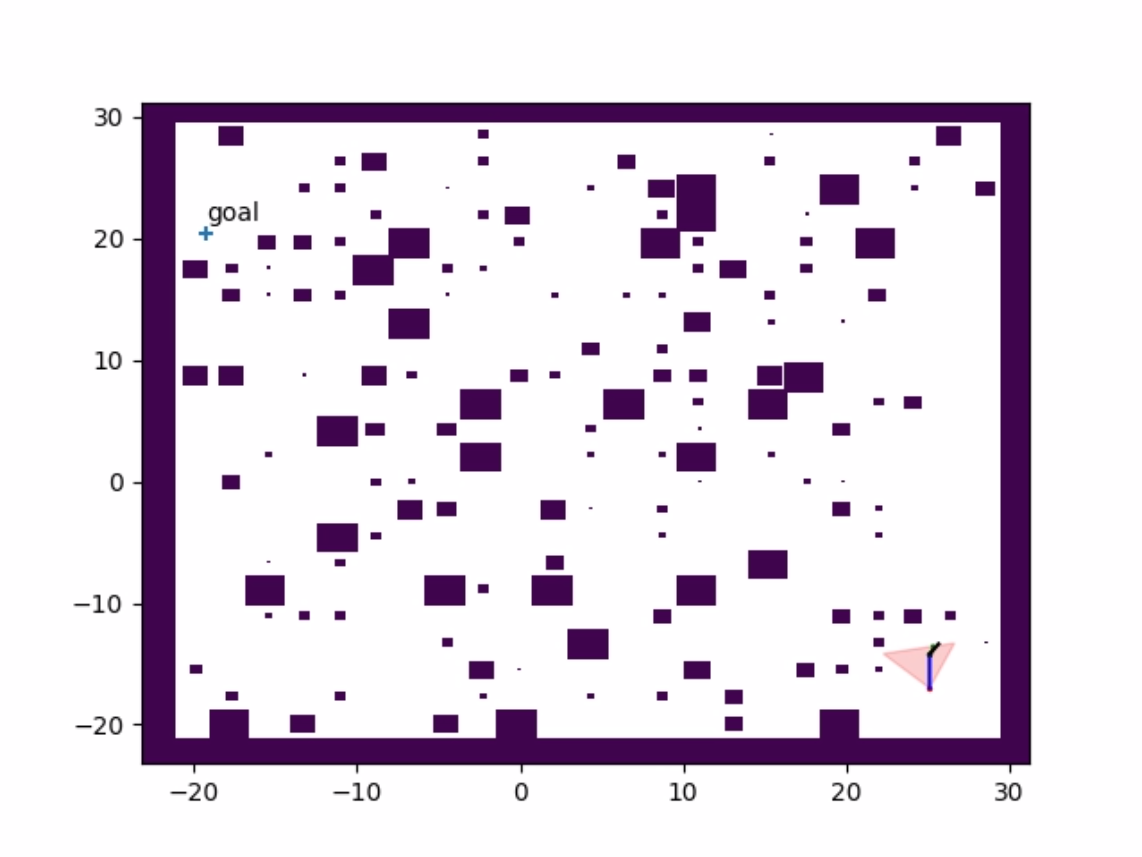}}\vspace{-5pt}\\

\caption{2D Simulation Environments}\label{fig:grid_maze}
\vspace{-0.025\textheight}

\end{figure*}

 We adopted the ellipsoidal intersection method discussed in 
\cite{ellipsoidintersect_Gilit_2014} for creating a sparsely connected graph using the intersections between the ellipsoids. We then augment the $\mathcal{X}_a$ and the global goal position as the start and end nodes in the graph. The shortest path to the global goal position is computed using Floyd–Warshal algorithm and the waypoints along the path are used as sequential goal points for the MPC. 

An illustration of sequential goal points forming the shortest path in the ellipsoidal free space can be seen in Fig \ref{fig:sequential_goal}. The agent and the target start from the bottom right of the environment. They traverse through the intersection points marked in magenta color and ultimately reach the goal position. When the agent reaches the ellipsoid which contains the final goal, the global planning is considered complete and we use the global goal position as $\mathcal{X}_g$ for the MPC.
\section{Experimental Results}\label{sec:results_and_discussion}

We robustly validated our proposed robot guided evacuation algorithm in a diverse range of simulated and real-world environments. The experimental setup for each of these environments is presented below.

\subsection{Experiment Setups}
We first evaluate the effectiveness of the proposed algorithm with environmental context as primitives for the non-linear MPC in simulated 2D Environments as shown in Fig. \ref{fig:grid_maze}. The hand-off guidance method in \cite{8956307} is re-implemented for a baseline comparison with our proposed approach. For the hand-off method, a fixed global path is used as a priori information and the path is split into multiple checkpoints, acting as guidelines for a simulated target to follow. The hand-off method assumes the target would always follow along the checkpoints to reach the safe location.
The 2D environment is classified into structured and unstructured environments. The structured environments shown in Fig. \ref{fig:grid_maze}(a) and \ref{fig:grid_maze}(b) represent common urban office space or indoor space. The unstructured environment shown in Fig. \ref{fig:grid_maze}(c) represents sparsely scattered obstacles of different shapes or sizes, thus representing a forest or outdoor space. 
Secondly, we simulate a large-scale robot-guided evacuation scenario, by controlling a virtual human avatar to follow a guiding UAV in Astralis 3D Simulator \cite{astralis_sim}. For the simulation, we reconstructed a real-world environment with random furniture obstacles blocking the passageways. 
Lastly, we implemented our proposed evacuation guiding algorithm in a real-world environment to study the perspective of a human evacuee following a guiding UAV and evaluating the performance of the algorithm in the presence of real-world noise. When the target is obstructed by the obstacle, the last tracked position would be used to re-adjust the guiding agent to re-establish line-of-sight with the target. The video of the real-world experiment can be viewed here: \url{https://youtu.be/XqMkWLGWwGg}.

\subsection{2D Grid Maze Validation of Evacuation Algorithm} \label{sec:gridmaze}
We evaluate our proposed guided evacuation algorithm with a target in a 2D grid maze as illustrated in Fig \ref{fig:grid_maze}. 
The grid maze covers a space of $54$m x $54$m. The agent guides a target in the simulation. In the hand-off method, the target would follow a given global path at \SI{1}{ms^{-1}}. For every \SI{0.1}{s} simulated time, a Gaussian noise of \SI{0.15}{m} is introduced for variabilities in the target's movement. For our proposed approach, the target has a $90^\circ$ FOV and \SI{7.5}{m} viewing distance. 
It moves at \SI{1}{ms^{-1}} speed toward the agent if the agent is \SI{1}{m} away and is in its view, otherwise the target remains stationary in position but constantly pivots its viewing angle to simulate scanning for the guiding agent.

In Eq. \ref{eqn:h_balance}, $\beta$ controls how steeply the agent should change its motion between guiding and going back to the human's view. $c$ determines the minimum acceptable distance to be away from the target before leading. 
We set $\beta$ and $c$ to be $4$ based on our empirical analysis.
With reference to these values of $\beta$ and $c$, such minimum acceptable distance is \SI{1}{m}. For the contextual weight in Eq. \ref{eqn:p_value}, we set $p_{j = \{1,..,4\}}$ to be $[0.05, 0.10, 0.15, 0.20]$. The MPC minimizes the combined distance cost functions with \SI{5}{s} horizon which consists of $40$ timesteps. 
The MPC solution is calculated using Casadi \cite{casadi_2018_Anderson} and IPOPT \cite{ipopt_2006_Wachter}. The calculation was done on a computer with an Intel® Core™ i7-7700K CPU @ 4.20GHz and has an average computation time of \SI{200}{ms}.

Each experiment is repeated $40$ times. The human and the agent are initialized near each other at $40$ different initial locations. These locations are randomly chosen in the obstacle-free space and have an average Euclidean distance of $45.5$ m from the fixed exit. We record the percentage of time in which the agent is visible to the human (Attention Rate) and the total movement time (TMT) in Table \ref{tab:grid_maze_result}. Due to the inherent assumption of the hand-off method, the target consistently follows checkpoints, thus, it is inappropriate to use Attention Rate as a comparative metric for this method.

We first compare our proposed algorithm against the baseline hand-off method. As highlighted above, the baseline method has a stringent requirement of full target attention on the guiding robot at all times. This is impractical in real-world rescue scenarios as the target may irrecoverably lose line-of-sight with the robot with increasing complexity of the environment. In contrast, our proposed approach maintains a high attention score regardless of the environment type. 
In addition, we observe an improvement in TMT for our proposed algorithm, particularly in the Simple (Fig. \ref{fig:grid_maze}(a)) and the Maze (Fig. \ref{fig:grid_maze}(b)) environments when contextual information is taken into account. However, our proposed method results in a longer TMT compared to the baseline hand-off method in the Forest Environment (Fig. \ref{fig:grid_maze}(c)). This is due to the increased complexity of the environment which results in several instances where line-of-sight with the target is disrupted requiring longer time for the guiding robot to retract to the last known target location. Thus, the overall total mission completion time is longer for the Forest Environment.
\begin{table}[hbt!]
\renewcommand{\baselinestretch}{1.4} 
\centering
\caption{Grid Maze Experiment Results}
\resizebox{1\columnwidth}{!}{\begin{tabular}{|lccccc|}
\hline
\multicolumn{1}{|c}{\textbf{}}                                                                      & \textbf{}                           & \textbf{\begin{tabular}[c]{@{}c@{}}Mean \\ Attention Rate \\ (\%)\end{tabular}} & \textbf{\begin{tabular}[c]{@{}c@{}}Median \\ Attention Rate\\ (\%)\end{tabular}} & \textbf{\begin{tabular}[c]{@{}c@{}}Mean \\ TMT\\ (s)\end{tabular}} & \textbf{\begin{tabular}[c]{@{}c@{}}Median \\ TMT \\ (s)\end{tabular}} \\ \hline
\multicolumn{1}{|l|}{\multirow{3}{*}{\begin{tabular}[c]{@{}l@{}}Simple\\ Environment\end{tabular}}} & \multicolumn{1}{l}{hand-off method} & -                                                                               & -                                                                                & \multicolumn{1}{l}{204.6}                                         & 203.7                                                                \\ \cline{2-6} 
\multicolumn{1}{|l|}{}                                                                              & without context                        & 86.7                                                                            & 90.3                                                                             & 181.5                                                              & 123.5                                                                 \\ \cline{2-6} 
\multicolumn{1}{|l|}{}                                                                              & with context                        & \textbf{87.4}                                                                   & 90.3                                                                             & \textbf{140.4}                                                     & \textbf{116.3}                                                        \\ \hline
\multicolumn{1}{|l|}{\multirow{3}{*}{\begin{tabular}[c]{@{}l@{}}Maze\\ Environment\end{tabular}}}   & \multicolumn{1}{l}{hand-off method} & -                                                                               & -                                                                                & \multicolumn{1}{l}{195.3}                                         & 194.7                                                                \\ \cline{2-6} 
\multicolumn{1}{|l|}{}                                                                              & without context                        & 88.1                                                                            & 87.6                                                                             & 209.3                                                              & 128.5                                                                 \\ \cline{2-6} 
\multicolumn{1}{|l|}{}                                                                              & with context                        & 87.3                                                                            & \textbf{92.5}                                                                    & \textbf{189.3}                                                     & \textbf{120.6}                                                        \\ \hline
\multicolumn{1}{|l|}{\multirow{3}{*}{\begin{tabular}[c]{@{}l@{}}Forest\\ Environment\end{tabular}}} & \multicolumn{1}{l}{hand-off method} & -                                                                               & -                                                                                & 274.3                                                              & 273.6                                                                 \\ \cline{2-6} 
\multicolumn{1}{|l|}{}                                                                              & without context                        & 84.5                                                                            & 88.6                                                                             & 570.6                                                              & 479.6                                                                 \\ \cline{2-6} 
\multicolumn{1}{|l|}{}                                                                              & with context                        & \textbf{87.9}                                                                   & \textbf{91.1}                                                                    & \textbf{542.2}                                                     & 518.0                                                                 \\ \hline
\end{tabular}}
\label{tab:grid_maze_result}

\end{table}

We compare the "Without Context" and the "With Context" versions of our work, using turning points as biases in keeping the guiding robot more visible to the target. The use of contextual information shows a consistent result for an immediate and smoother follow-up by the target, especially near the turns. As a result, we see a consistent reduction in TMT for the Simple and Maze Environments ($22.6\%$ and $9.6\%$), respectively. In the case of the Forest Environment (Fig. \ref{fig:grid_maze}(c)), there is an increase in median TMT. By defining contextual turning points in the Forest Environment, there is a high probability of multiple turning points within a short traversing distance. Hence, our agents are biased towards maintaining visibility rather than guiding towards the goal location. As a result, the pair moves slower to reach the goal location, as reflected by the higher median TMT.

\subsection{3D Simulation Validation for Evacuation Algorithm}
\begin{figure}[htb]
\centering
\includegraphics[width=\columnwidth]{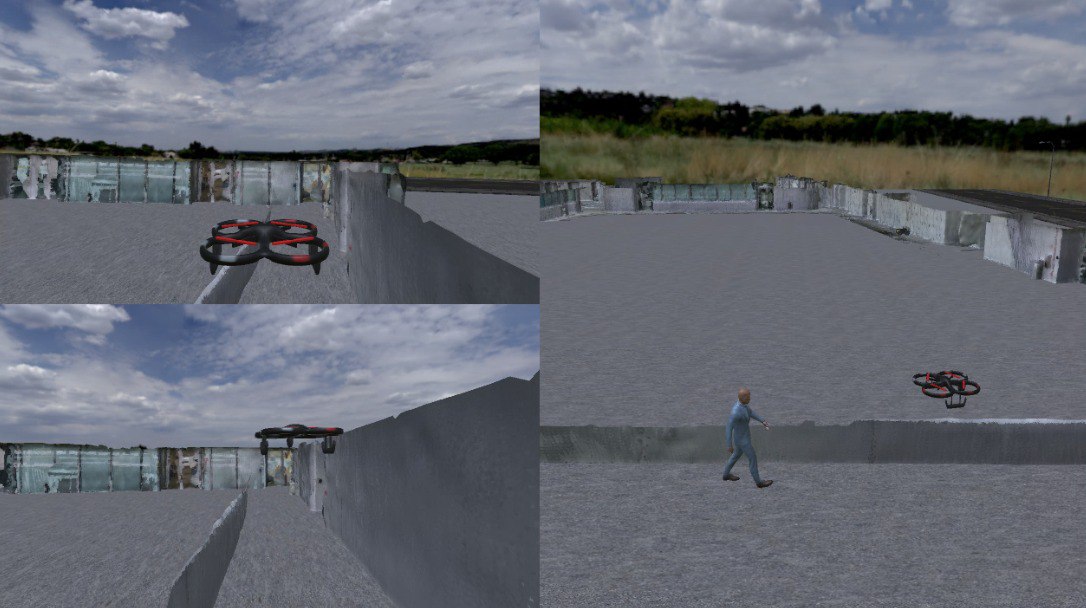}
\caption{\label{fig:scanned_building}
UAV guiding human avatar in 3D scanned environment: the top left image shows the third-person view from behind the UAV, the bottom left image shows the human target's view, and the right image shows the pair's side view.}
\vspace{-0.01\textheight}
\end{figure}

\begin{figure}[htb]
\vspace{0.01\textheight}
\centering
\includegraphics[width=0.95\columnwidth]{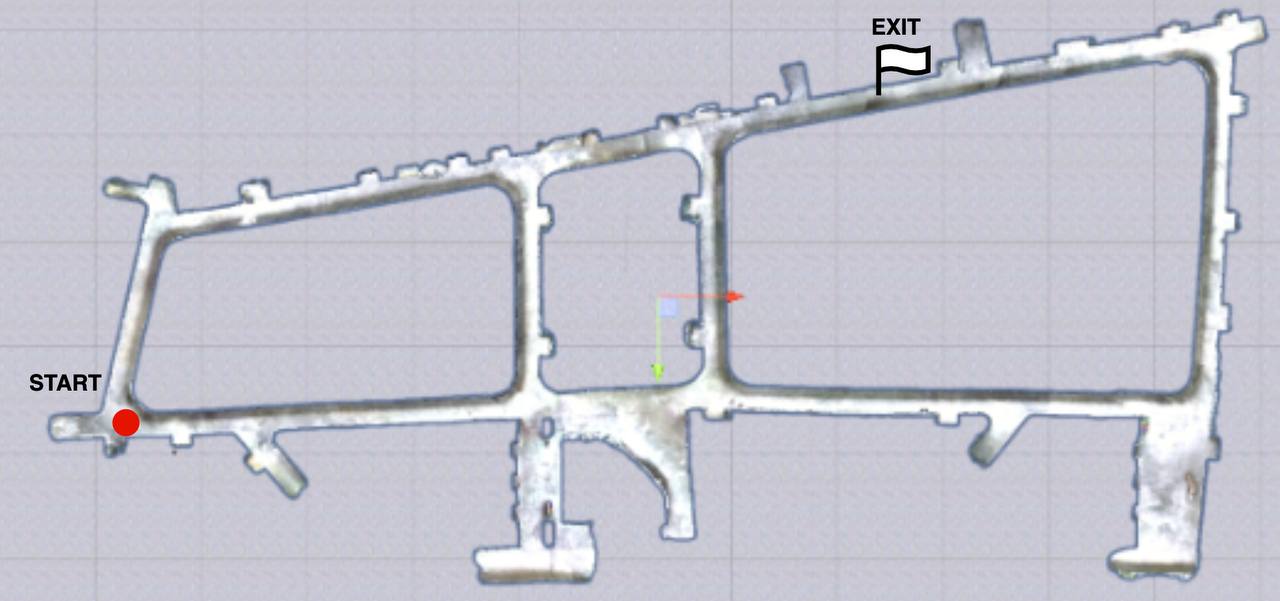}
\caption{\label{fig:leve_seven_topdown} Top-down view of the simulated campus environment }
\vspace{-0.015\textheight}
\end{figure}

We integrate the MPC module into ROS \cite{ros} and recreated a 3D model of our university's campus floor (\SI{110}{m} x \SI{40}{m}) by using the photogrammetry import feature from Astralis 3D simulator \cite{astralis_sim}.  A virtual human avatar is the target and it is controlled by a real human user to cooperatively follow the guiding UAV toward the goal location as shown in Fig. \ref{fig:scanned_building}. The human avatar has a FOV $70^{\circ}$ and moves at \SI{1}{ms^{-1}}. We only use the first-person camera view from the human avatar as illustrated by the bottom left image of Fig. \ref{fig:scanned_building} to control the human avatar for following the guiding UAV. We assume the human avatar is fully cooperative in following the UAV.
The top-down view of the environment is shown in Fig. \ref{fig:leve_seven_topdown}, where the pair are initialized at the bottom left and the exit location is marked by the flag symbol. There are four pathways in which the UAV can guide the human target to the goal location. 
We validate the proposed algorithm with three environment configurations, namely: zero, medium and high messiness using Astralis 3D simulator. For zero messiness, there are no added obstacles in the environment. For medium messiness, there are some obstacles but none that completely block the pathways. These obstacles could block the view of the human avatar. For high messiness, two pathways are completely blocked by obstacles. For each environment configuration, we executed $3$ iterations of the algorithms. The averaged TMT is given in Table \ref{tab:rset_eval}.

\begin{table}[hbt]
\renewcommand{\baselinestretch}{2} 
\centering
\caption{Mean TMT Based on Environment Complexities}

\resizebox{1\columnwidth}{!}{\begin{tabular}{|cccc|}
\hline
                      & \textbf{Zero Messiness} & \textbf{Medium Messiness} & \textbf{High Messiness} \\ \hline
\textbf{Without Context} & 167.0 s                 & 269.2 s                   & 470.7 s                 \\ \hline
\textbf{With Context} & 188.7 s                 & 222.7 s                   & 321.2 s                 \\ \hline
\end{tabular}}

\label{tab:rset_eval}
\end{table}

The observed results in Table \ref{tab:rset_eval} support our hypothesis of using environmental context to allow better guidance behavior in evacuation. When there is zero messiness (no path blockage and no view-blocking obstacles), the TMT is the least. However, when comparing "with context" scenario against "without context" we observe an increase in TMT. This is because, given the contextual information, the guiding UAV slows down at intersections to ensure visibility to the human target. 
In the case of medium and high messiness, we see a significant reduction in TMT for "with context" scenarios compared to "without context" ($13.3\%$ and $31.8\%$) as the slower speed of the robot in "with context" scenarios helps in guiding the human efficiently. Thus, when the environment becomes more uncertain, the "without context" version would have more instances where the guiding robot goes beyond the field of view of the human, thus would have to spend extra time to come back to human's view to resume the guiding action.  
We would like to highlight that our evacuation guiding algorithm (Sec. \ref{sec:evacuation_algo}) was only implemented in 2D, hence the guiding UAV are programmed to fly at a fixed height in 3D environment. The 3D obstacles are also flattened into 2D, which may sometimes result in ineffective robot navigation. 

\subsection{Real-World Validation of Evacuation Algorithm}
The physical experiment analyzes the impact of using contextual information by assessing the inter-agent distance and the motion of the human follower. The experiment was performed at the Singapore University of Technology and Design (SUTD), specifically in an L-shaped turn corridor. The test environment occupies an area of \SI{5}{m} x \SI{24}{m}, as depicted in Fig \ref{fig:floor_plan}. At the start of each round of the experiment, the human evacuee is positioned at the orange dot (shaded). The objective is for the human to follow the robot and reach the exit located at the end of the turn in the corridor, represented by a green dot (not shaded). Since there is only one type of turn in the environment, the contextual weight in Eq. \ref{eqn:p_value} is fixed at $0.1$. Total $4$ sets of trajectories from the same day of the experiment are used for comparison on the proposed algorithm with and without context cost.
\begin{figure}[hbt!]
\centering
\subfigure[Without Context]{\includegraphics[width=0.46\columnwidth]{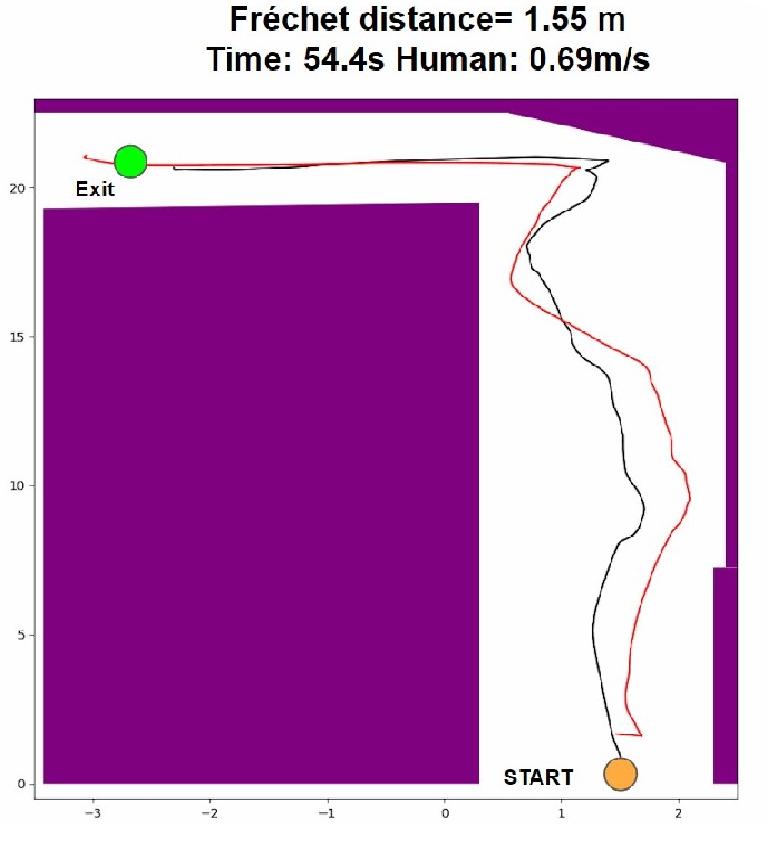}}
\subfigure[With Context]{\includegraphics[width=0.46\columnwidth]{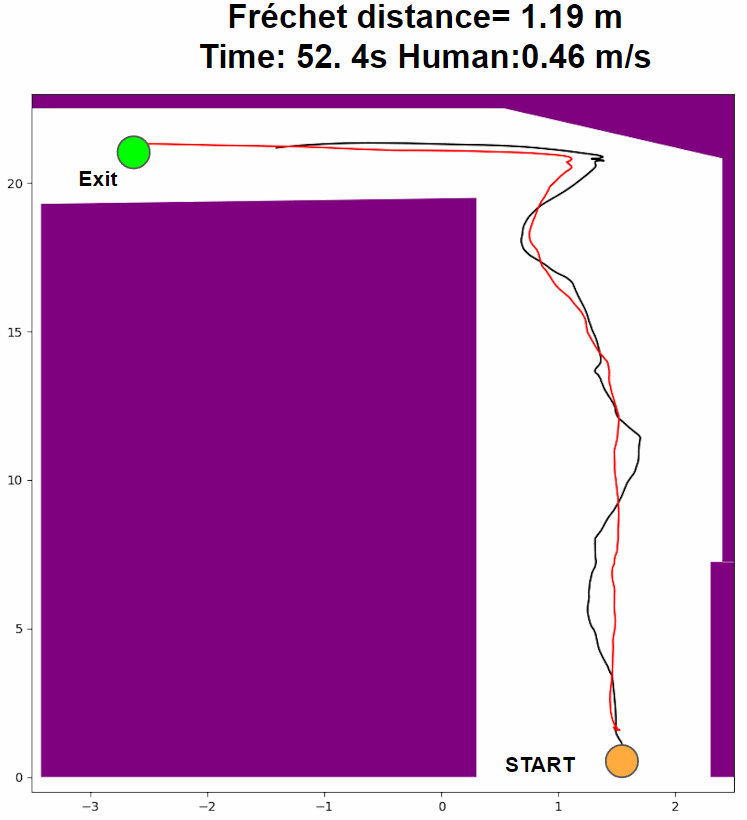}}
\caption{Trajectory comparison for performing evacuation guidance in the L-shape corridor. The human's trajectories are in black, the guiding agent's trajectories are in red, and the obstacles are in purple. }\label{fig:human_follower_trajecotry}
\vspace{-0.01\textheight}
\end{figure} 

The experiment utilizes the Local Positioning System (LPS) \cite{bitcrazy_lps} from Crazyflie. The set-up consists of two agents: a guiding Crazyflie (leader) and a human-position update Crazyflie (follower) mounted on the hat or hand-held by the human subject. The Crazyflie continuously provides 6-DoF pose information at 10 Hz. By mounting the device on the human follower's hat or holding the device facing the forward direction of the human follower, we can directly capture real-time updates of the human follower's position and orientation.
Fig \ref{fig:human_follower_trajecotry} illustrates an example of a trajectory plot of with and without contextual cost. Given the LPS system has a localization error of $\pm$\SI{10}{cm}, the noisy trajectories are smoothened by interpolating a 5-second moving average window. 
We noted that the human observes the movements of the guiding drone, but may not precisely replicate the exact path taken by the drone. Moreover, there are temporal delays when the human responds to changes in the guiding drone's motion, especially nearing the turn. To account for these spatial and temporal differences, Fréchet distance is employed as a way to assess the similarity between the human follower and the leading drone. Additionally, Fréchet distance helps to analyze whether the desired inter-agent distance is consistently maintained.
In order to compare our proposed context-aware evacuation guidance algorithm and the baseline "without context" algorithm, we perform statistical significance test on the raw data for the respective Fréchet distances and target speed.
\begin{figure}[t]
\centering
\subfigure[Fréchet Distance Box Plot\vspace{-0.03\textheight}]{\includegraphics[width=0.90\columnwidth]{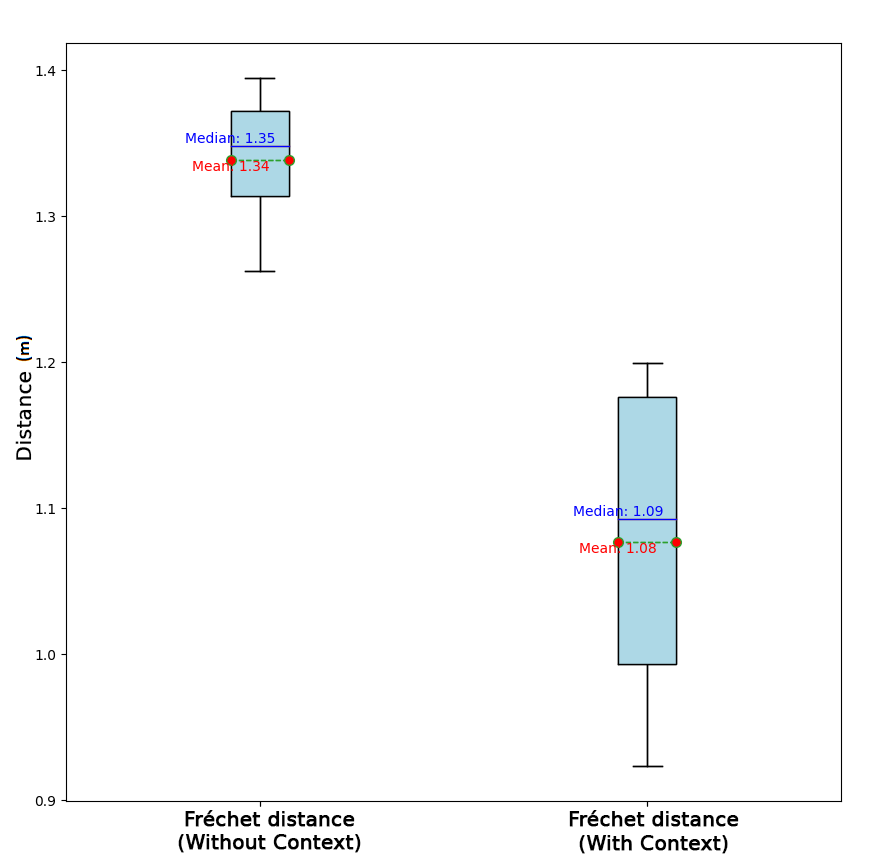}\label{fig:box_distance}}
\subfigure[Speed Profile Box Plot]{\includegraphics[width=0.90\columnwidth]{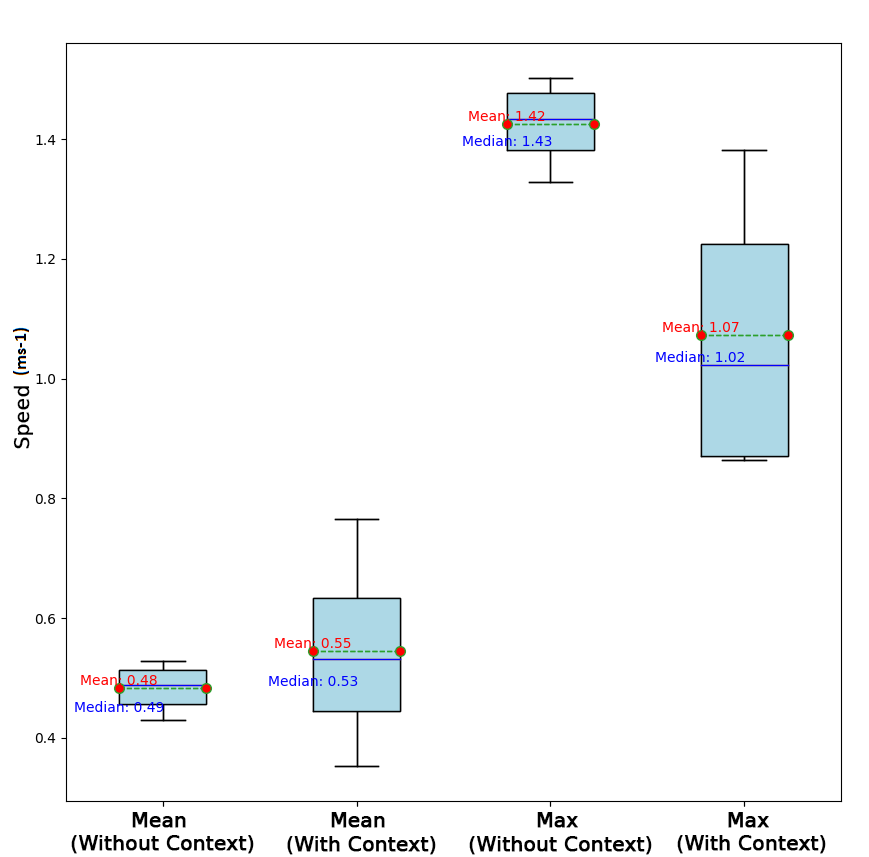}\label{fig:box_speed}}
\caption{Box plot comparison for the speed profile and the Fréchet distance for baseline MPC algorithm and the "with context version".}\label{fig:boxplot}
\vspace{-0.015\textheight}
\end{figure} 

In Fig \ref{fig:human_follower_trajecotry}(b), the effect of introducing a context cost of $0.1$ is observed which brings the guiding robot closer to the human as they move together. This observation is reinforced by the Fréchet distance box plot in Fig \ref{fig:box_distance}, which reports mean and median Fréchet distances of \SI{1.34}{m} and \SI{1.35}{m} for the "without-context", and \SI{1.08}{m} and \SI{1.09}{m} for the "with context". The difference between the two sets of Fréchet distance data is statistically significant based on one-way ANOVA testing ($F=13.6, p<0.05$). Post-hoc Tukey's HSD tests revealed that the Fréchet distances in the "with context" are significantly smaller than the Fréchet distances in the "without-context" ($F=13.6, p<0.02$).
A $19.4\%$ reduction in average  Fréchet distance indicates that the human is able to follow more closely with the guiding UAV, and their paths become more similar. This aligns with the initial hypothesis that the introduction of a context cost of $0.1$ enables the human to move closer to the drone and follow its path more consistently.

The shorter Fréchet distance suggests that the guiding robot and the human, as a pair, move at a slower pace as they approach the goal location. This adjustment can be interpreted as a way to maintain visibility. The maximum speed of the human is used to represent the pacing of the human evacuee. The speed box plot in Fig. \ref{fig:box_speed} reports mean and median of maximum speeds as \SI{1.42}{ms^{-1}} and \SI{1.43}{ms^{-1}} for the "without context", and \SI{1.07}{ms^{-1}} and \SI{1.02}{ms^{-1}} for the "with context". The difference between the two sets of maximum speed data is statistically significant based on one-way ANOVA testing ($F=7.18, p<0.05$). Post-hoc Tukey's HSD tests revealed that the maximum human speed in the "with context" is slower than the maximum human speed in the "without context" ($F=2.68, p<0.065$). A reduction of $24.6\%$ in maximum human speed further indicates that slower robot speed is required to synchronize with the human.

\section{Conclusion and Future Work}
This paper presents an active robot guidance algorithm for evacuation scenarios that utilize viewpoint constraints and environmental context as primitives in a non-linear model predictive control (MPC) framework. The algorithm has been successfully implemented on physical robot platforms, generating trajectories that ensure collision-free navigation for the robot while enabling the target to follow. The experimental results demonstrate that incorporating environmental context leads to improved visibility and shorter total evacuation time. For the human evacuee's experiment, the human follower could follow the leading drone more consistently when context information is considered. In addition, in complex unstructured environments, our proposed approach consistently maintains significant attention time to the target with a trade-off on the total movement time. The increased attention time being one of the most important features required for a practical rescue mission. In our future work, we will also aim to reduce the overall mission time by using more generic contextual information.

\section*{Acknowledgement}
This research was supported by the National Research Foundation, Prime Minister’s Office, Singapore, under its AI Singapore Program (AISG Award No: AISG2-RP-2020-016).
\bibliographystyle{unsrt}
\bibliography{IROS24_Robot_Evacuation}

\end{document}